\documentclass[sigconf]{acmart}
\makeatletter
\def\@ACM@checkaffil{
    \if@ACM@instpresent\else
    \ClassWarningNoLine{\@classname}{No institution present for an affiliation}%
    \fi
    \if@ACM@citypresent\else
    \ClassWarningNoLine{\@classname}{No city present for an affiliation}%
    \fi
    \if@ACM@countrypresent\else
        \ClassWarningNoLine{\@classname}{No country present for an affiliation}%
    \fi
}
\makeatother

\AtBeginDocument{%
  \providecommand\BibTeX{{%
    \normalfont B\kern-0.5em{\scshape i\kern-0.25em b}\kern-0.8em\TeX}}}

\acmConference[KDD '23]{Make sure to enter the correct
  conference title from your rights confirmation emai}{June 03--05,
  2018}{Woodstock, NY}
\acmPrice{15.00}
\acmISBN{978-1-4503-XXXX-X/18/06}

\usepackage{booktabs, multirow} 
\usepackage{soul}
\usepackage{balance}
\usepackage{enumitem}
\usepackage{todonotes}
\newcommand{\model}{\texttt{mTLDRgen}}
\newcommand{\data}{\texttt{mTLDR}}
\usepackage{amsmath}
\usepackage{amsfonts}
\newcommand{\R}{\mathbb{R}}

\copyrightyear{2023}
\acmYear{2023}
\setcopyright{acmlicensed}\acmConference[KDD '23]{Proceedings of the 29th ACM SIGKDD Conference on Knowledge Discovery and Data Mining}{August 6--10, 2023}{Long Beach, CA, USA}
\acmBooktitle{Proceedings of the 29th ACM SIGKDD Conference on Knowledge Discovery and Data Mining (KDD '23), August 6--10, 2023, Long Beach, CA, USA}
\acmPrice{15.00}
\acmDOI{10.1145/3580305.3599830}
\acmISBN{979-8-4007-0103-0/23/08}

\begin{document}

\title{Fusing Multimodal Signals on Hyper-complex Space for Extreme Abstractive Text Summarization (TL;DR) of Scientific Contents}

\author{Yash Kumar Atri}
\email{yashk@iiitd.ac.in}
\affiliation{%
  \institution{IIIT Delhi}
}

\author{Vikram Goyal}
\email{vikram@iiitd.ac.in}
\affiliation{%
  \institution{IIIT Delhi}
}

\author{Tanmoy Chakraborty}
\email{tanchak@iitd.ac.in}
\affiliation{%
  \institution{IIT Delhi}
}

\renewcommand{\shortauthors}{Yash Kumar Atri, Vikram Goyal, \& Tanmoy Chakraborty}


\begin{abstract}
The realm of  scientific text summarization has experienced remarkable progress due to the availability of annotated brief summaries and ample data. However, the utilization of multiple input modalities, such as videos and audio, has yet to be thoroughly explored. At present, scientific multimodal-input-based text summarization systems tend to employ longer target summaries like abstracts, leading to an underwhelming performance in the task of text summarization.

In this paper, we deal with a novel task of {\em extreme abstractive text summarization} ({\em aka} {\em TL;DR generation}) {\em by leveraging multiple input modalities}. To this end, we introduce \data, a first-of-its-kind  dataset for the aforementioned task, comprising videos, audio, and text, along with both author-composed summaries and expert-annotated summaries. The \data\ dataset accompanies a total of $4,182$  instances collected from various academic conference proceedings, such as ICLR, ACL, and CVPR. Subsequently, we present \model, an encoder-decoder-based model that employs a novel dual-fused hyper-complex Transformer combined with a Wasserstein Riemannian Encoder Transformer, to dexterously capture the intricacies between different modalities in a hyper-complex latent geometric space. The hyper-complex Transformer captures the intrinsic properties between the modalities, while the Wasserstein Riemannian Encoder Transformer captures the latent structure of the modalities in the latent space geometry, thereby enabling the model to produce diverse sentences. \model\ outperforms 20 baselines on \data\ as well as another non-scientific dataset (How2) across three Rouge-based evaluation measures.
Furthermore, based on the qualitative metrics, BERTScore and FEQA, and human evaluations, we  demonstrate that the summaries generated by \model\ are fluent and congruent to the original source material.

\end{abstract}





\maketitle

\begin{figure}[!t]
    \centering
    \scalebox{0.47}{
    \includegraphics[trim={0.5cm 1.5cm 6.5cm 0cm},clip]{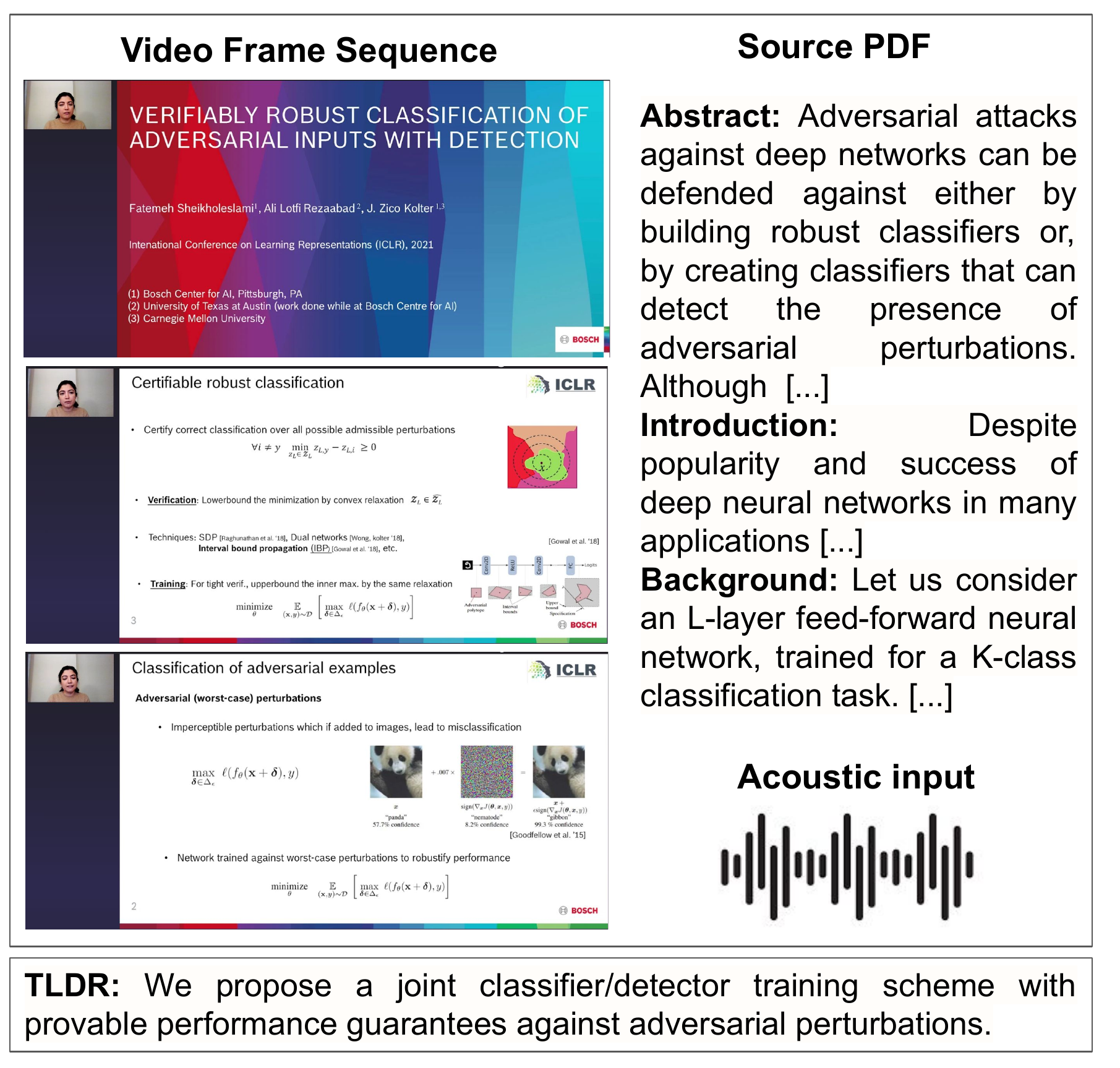}}
    \caption{A sample of \data\ dataset with video, text and audio modalities along with the target TLDR. The feature representations for video frames are obtained by ResNext, audio features are extracted using Kaldi, and the text is extracted from the pdf of the article.}
    \label{fig:sampleexample}
    \vspace{-3mm}
\end{figure}

\section{Introduction}
Abstractive text summarization enables one to promptly comprehend the essence of a written work, determining if it is worth perusing. In contrast to extractive summarization, which emphasizes the crucial passages within the original document as a summary, abstractive summarization recomposes the summary from scratch by synthesizing the core semantics and the entire substance of the document. 
Earlier studies dealt with abstractive summarization by solely utilizing textual input \citep{see2017get, lebanoff-song-liu:2018, gehrmann2018bottom, ijcai2020-0514, CHUNG2020105363}; thereafter, multimodal inputs \citep{shah2016leveraging, li2018read, zhu-etal-2018-msmo, palaskar2019multimodal} were integrated to enhance the quality of the generated summaries. Studies have revealed that multimodal data assists humans in comprehending the essence of a written work more effectively \cite{GIANNAKOS2019108}, thus leading us to the inference that multimodal data can enrich the context and produce more comprehensive scientific summaries.

{\bf Motivation:} With the emergence of deep learning architectures like LSTM, Attention, and Transformer, the literature in the scientific community has skyrocketed. It is extremely hard to keep up with the current literature by going through every piece of text in a research article. The abstract of a paper often serves as a bird's eye view of the paper, highlighting the problem statement, datasets, proposed methodology, analysis, etc. Recent studies \cite{ATRI2021107152} re-purpose  abstracts to generate summaries of scientific articles. However, it is cumbersome to go through the abstract of each paper. The abstracts are nearly $300$ tokens long, and reading the complete abstract of every paper to figure out the mutual alignment is tedious. The task of TL;DR ({\em aka}, tl;dr, too long; didn't read) \cite{voelske:2017,cachola-etal-2020-tldr} was introduced to generate an extremely concise summary from the text-only article highlighting just the high-level contributions of the work. Later, \citet{citesum} introduced the CiteSum dataset for generating text-only extreme summaries. However, the text alone can not comprehend the entire gist of the research article. The multimodal information, including the video of the presentation and audio, often provide crucial signals for extreme text summary generation. 

{\bf Problem statement:} In this work, we propose a new task of multimodal-input-based  TL;DR generation for scientific contents which aims to generate an extremely-concise and informative text summary. We incorporate the visual modality to capture the visual elements, the audio modality to capture the tonal-specific details of the presenter, and the text modality to help the model align all three modalities. We also show the generalizability of the proposed model on another non-academic dataset (How2).

{\bf State-of-the-art and limitations:} The pursuit of multimodal-input-based abstractive text summarization can be related to various other fields, such as image and video captioning \citep{mun2016text, liu2020sibnet, iashin2020multi, shi2020video, shen2020remote}, video story generation \citep{gella2018dataset}, video title generation \citep{zeng2016generation}, and multimodal sentence summarization \citep{li2018multi}. However, these works generally produce summaries based on either images or short videos, and the target summaries are easier to predict due to the limited vocabulary diversity. On the other hand, scientific documents have a complex and structured vocabulary, which the existing methods \cite{palaskar2019multimodal} of generating short summaries are not equipped to handle. Recently,  \citet{ATRI2021107152}  proposed as a novel dataset for the multimodal text summarization of scientific presentations; however, it uses the abstract as the target summary, which falls short in producing coherent summaries for the extreme multimodal summarization (TL;DR) task.

In summary, the current paper offers the following contributions:
\begin{itemize}[noitemsep, nolistsep, leftmargin=1em]
    \item {\bf  Novel problem:} We propose the task of extreme abstractive text summarization for scientific contents, by utilizing videos, audio and research articles as inputs.
    \item {\bf  Novel dataset:} The development and curation of the first large-scale dataset \data\ for extreme multimodal-input-based text summarization of scientific contents. Figure \ref{fig:sampleexample} shows an excerpt from the \data\ dataset. This dataset has been meticulously compiled from five distinct public websites and comprises articles and videos obtained from renowned international conferences in Computer Science. The target summaries are a fusion of manually-annotated summaries and summaries written by the authors/presenters of the papers.
    
    \item {\bf  Novel model:} We propose {\model}, a novel encoder-decoder-based model designed to effectively capture the dynamic interplay between various modalities. The model is implemented with a dual-fused hyper-complex Transformer and a Wasserstein Riemannian Encoder Transformer. The hyper-complex Transformer projects the modalities into a four-dimensional space consisting of one real component and three imaginary components, thereby capturing the intrinsic properties of individual modalities and their relationships with one another. Additionally, the Wasserstein Riemannian Encoder Transformer is employed to apprehend the latent structure of the modalities in the geometry of the latent space.
    
    \item {\bf Evaluation:} We benchmark \data\ over six extractive (text-only), eight abstractive (text-only), two video-based and four multimodal summarization baselines, demonstrating the effectiveness of incorporating multimodal signals in providing more context and generating more fluent and informative summaries. We evaluate the benchmark results over the quantitative (Rouge-1/2/L) and qualitative (BERTScore and FEQA) metrics. Our proposed modal, \model, beats the best-performing baseline by $+5.24$ Rouge-1 and $+3.35$ Rouge-L points. We also show the generalizability of \model\ on another non-scientific dataset (How2).

    \item{\bf Deployment:} We further designed an in-house institute-wide web API based on the end-to-end pipeline of \model. The web API is currently undergoing a beta testing phase and has gathered more than $100+$ hits so far. The API will be made open across academic institutes and beyond upon successful completion of the beta testing. 
    \end{itemize}
    
    {\bf Reproducibility:} We discuss the detailed hyperparameters (Supplementary, Table \ref{tab:params}) and experimentation setting in Section \ref{sec:training_params}. We also provide  a sample dataset of \data\ and the source code of \model\ at \url{https://github.com/LCS2-IIITD/mTLDRgen}.

\section{Related Work} 
The development and utilization of abstractive text summarization systems involve the formulation of textual summaries through the integration of two or more auxiliary signals. These signals, beyond the traditional text, may encompass video \cite{8664480}, images \cite{10.1007/978-3-031-20059-5_37}, and audio \cite{GonzalezGallardo2020AudioSW}. The integration of  additional modalities, as compared to text-only systems, offers a plethora of opportunities to enhance the contextual richness and knowledge base of the generated summaries. Several recent studies \cite{sanabria2018how2, ATRI2021107152} demonstrated that the integration of multimodal signals such as video and audio can significantly improve the contextual accuracy and informativeness of the summaries generated by unimodal systems.

\textbf{Unimodal text summarization:} Text summarization is classified into two categories -- extractive and abstractive. Extractive systems extract the most relevant sentences from the source document to form the summary, while abstractive systems paraphrase important sentences to generate a new summary. Conventional extractive summarization approaches either construct a graph representation of the source document \cite{Erkan_2004, textrank2004, 10.1145/3404835.3463111} or pose the summarization task as a binary classification with ranking \cite{nallapati2016summarunner, liu2019fine, cheng2016neural, zheng2019sentence}. On the other hand, abstractive summarization has significantly benefited from the advent of deep neural networks. Early works \cite{nallapati2016abstractive, see2017get} utilized CNN/Dailymail dataset \cite{hermann2015teaching} to explore abstractive summarization on a large scale. Later, Pointer Generators (PG) \cite{see2017get} were extended to capture the latent structures of documents \cite{song2018structure,Multinews2019}. The use of Transformers \cite{vaswani2017attention} and attention mechanisms \cite{DBLP:journals/corr/BahdanauCB14} further improved the encoding of long sequential data. These improvements include leveraging Transformers \cite{10.1007/978-981-16-9012-9_21} and repurposing attention heads as copy pointers \cite{gehrmann2018bottom} to enhance the quantitative performance. Large language models \cite{devlin2018bert,zhang2020pegasus,2020t5} have demonstrated impressive performance on multiple datasets \cite{hermann2015teaching, Multinews2019}. Models proposed in \citep{devlin2018bert} and \cite{zhang2020pegasus} are pre-trained using token and phrase masking techniques, respectively, while \citet{2020t5} approached all downstream tasks as a text-to-text problem and pre-train using a single loss function.

\textbf{Extreme unimodal text summarization:} The objective of extreme text summarization is to drastically reduce the size of the source document while preserving its essence in the resulting summary. The concept of extreme summarization was first introduced by \citet{voelske:2017} with a novel dataset focused on social media summarization. Subsequently, \citet{cachola-etal-2020-tldr} and \citet{citesum} presented new corpora, namely SciTLDR and CiteSum, respectively, for extreme summarization of scientific documents. However, it remains an open area to explore extreme abstractive text summarization using multimodal signals.

\textbf{Text summarization with multimodality:} The incorporation of multimodal information in text summarization enriches the comprehension of the data, elevating it to a representation that better reflects the source document \citep{kiela2017deep, baroni2016grounding}. In the absence of multimodal information, the summarization model can only comprehend limited information; however, with the integration of multiple modalities, the models acquire a more comprehensive understanding, leading to the creation of highly fluent and semantically meaningful summaries. Multimodal summarization has been explored in various domains, including instructional YouTube videos \cite{sanabria2018how2}, news videos \cite{li2016multimedia, chen2018abstractive}, and recipe videos \cite{zhou2017towards}. The concept of generating multimodal outputs from multimodal inputs has also been studied \cite{zhu-etal-2018-msmo, https://doi.org/10.48550/arxiv.2109.05812, https://doi.org/10.48550/arxiv.2204.03734}, where a news event was summarized with a text summary along with a corresponding image. \citet{https://doi.org/10.48550/arxiv.2108.05123} recently introduced cross-modal alignment to harmonize visual features with the text to produce a more coherent summary. Although existing datasets feature the use of either images or videos as a visual modality, the closest dataset to our task is the How2 dataset \cite{sanabria2018how2}, housing all three modalities to generate a short summary. However, when compared to \data, the How2 dataset falls short in terms of length, structuredness and complexity of vocabulary in the source and target documents. Evidently, the existing approaches over the How2 dataset fail to extend similar performance on the \data\ dataset. 

\begin{table}[!t]\centering
\caption{Statistics of the used datasets (\data\ and How2) -- the number of samples (\#source), average token length of source documents (avg source len), average tokens in the target summaries (avg target len), and abstractness percentage (Abs) of datasets.}\label{tab:datasetstats}

\begin{tabular}{lccccc}\toprule
Dataset &\#source &avg source len &avg target len &\%Abs \\\midrule
How2 &73993 &291 &33 & 14.2 \\
\data &4182 &5K &18 &15.9 \\
\bottomrule
\end{tabular}
\vspace{-4mm}
\end{table}

\section{Proposed Dataset}
To explore the efficacy of multimodal signals and enable enriched abstractive summaries aided by various modalities, we introduce {\bf \data}, the first large-scale  {\bf m}ultimodal-input based abstractive summarization ({\bf TL;DR}) dataset with diverse lengths of videos. \data\ is collected from various well-known academic conferences like ACL, ICLR, CVPR, etc. The only comparable dataset to \data\ is the How2 dataset, which comprises short instructional videos from various topics like gardening, yoga, sports, etc. Compared to How2, \data\ contains structured and complex vocabulary, which requires attention to diverse information while generating summaries.

Our compilation encompasses video recordings from openreview.net and videolecture.net, in addition to the accompanying source pdf and metadata information, including the details of the authors, title, and keywords. The collected dataset comprises a total of $4,182$ video recordings, spanning a duration of over $1,300$ hours. Of these, we designated $2,927$ instances as the training set, $418$ for validation, and $837$ for testing. The average length of the videos is $14$ minutes, and the TLDR summary has an average of $19$ tokens. The target summaries for the data are a combination of human-annotated and author-generated summaries. \textcolor{black}{In terms of abstractness, \data\ contains 15.9\% novel $n$-grams.} Each data instance includes a video, audio extracted from the video, an article pdf, and a target summary. We opted not to annotate or retain multiple summaries for a single instance to ensure efficient training and testing processes. We assert that a single extreme summary is sufficient to convey the essence of the paper. The target summaries for papers obtained from the ACL anthology were annotated as they lacked any author-generated summaries. Of the $4,182$ videos, a total of $1,128$ summaries were manually annotated by $25$ annotators. During the annotation process, the annotators were instructed to thoroughly read the abstract, introduction, and conclusion and to have a general understanding of the remaining content. Each summary was then verified by another to confirm that it accurately represents the paper's major contributions. 

In contrast, the How2 dataset \cite{sanabria2018how2} consists of $73,993$ training, $2,965$ validation, and $2,156$ test instances. The average token length for the source documents is $291$, while for the target summary, it is $33$. \textcolor{black}{Compared to the source document, the target summaries contains $14.2\%$ novel $n$-grams.} The transcripts for videos and the target summary are human-annotated. Table \ref{tab:datasetstats} shows brief statistics of the How2 and  \data\ datasets.


\section{Proposed Methodology}

\begin{figure*}[!t]
    \centering
    \scalebox{0.85}{
    \includegraphics[trim={0 0.1cm 0cm 0cm},clip]{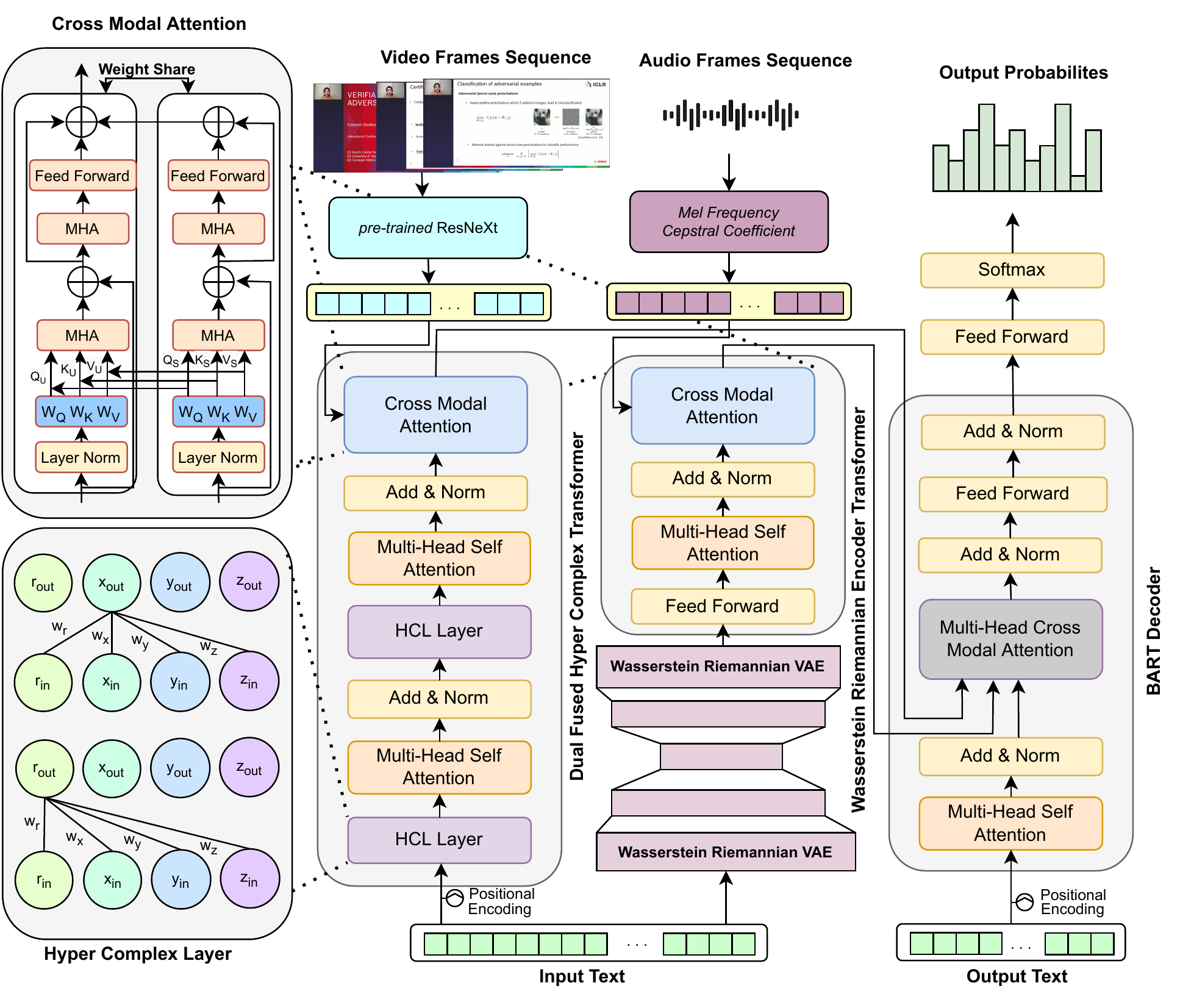}}
    \caption{An overview of the proposed model -- \model. It houses two parallel encoders, one with a hyper-complex layer fused with the video embeddings using cross-model attention and the other with Wasserstein Riemannian Encoder Transformer with audio embeddings fused with cross-model attention. The individual encoder representations are later fused with the multi-head attention of the pre-trained BART decoder to generate the final summary.}
    \label{fig:modelArch}
    \vspace{-4mm}
\end{figure*}

This section presents our proposed system, {\bf \model}, a {\bf m}ultimodal-input-based extreme abstractive text summary ({\bf TL;DR}) {\bf gen}erator. Figure \ref{fig:modelArch} shows a schematic diagram. During an academic conference presentation, there are typically three major modalities present -- visual, audio, and text, each of which complements the others, and when combined, contributes to a rich and expressive feature space, leading to the generation of coherent and fluent summaries. \model\ initially extracts features from the independent modalities and then feeds them to the dual-fused hyper-complex Transformer (DFHC) and the Wasserstein Riemannian Encoder Transformer (WRET) blocks. Cross-modal attention is used to fuse the visual and audio features with the text representations. Finally, the fused representation is fed to a pre-trained BART \cite{lewis2019bart} decoder block to produce the final summary. The rest of this section delves into the individual components of \model.

\subsection{Video Feature Extraction}
The video modality in an academic presentation often comprises variations in frames and kinesthetic signals,  highlighting key phrases or concepts during a presentation. To capture visual and kinesthetic aspects, we utilise the ResNeXt-152-3D \cite{kay2017kinetics} model as it is pre-trained on the Kinetics dataset for recognition of $400$ human actions. Four frames per second are extracted from the video, cropped to $112$ $\times$ $112$ pixels and normalized, and a 2048-dimensional feature vector is extracted from the ResNeXt-152-3D model for every 12 non-overlapping frames. \textcolor{black}{The 2048-dimensional vector is then fed to the mean pooling layer to obtain a global representation of the video modality. Later, a feed-forward layer is applied to map the 2048-dimensional vector to a 512-dimensional vector.}

\subsection{Speech Feature Extraction}
To capture the variations in the speaker's voice amplitudes, which are considered to signify the importance of specific topics or phrases \cite{tepperman2006yeah}, we extract audio features from the conference video. This is accomplished by extracting audio from the video using the FFMPEG package\footnote{https://ffmpeg.org/}, resampling it to a mono channel, processing it to a $16K$ Hz audio sample, and dividing it into overlapping windows of $30$ milliseconds. The extracted audio is then processed to obtain $512$-dimensional Mel Frequency Cepstral Coefficients (MFCC) features. The final representation is obtained by applying a log Mel frequency filter bank and discrete cosine transformation, and the feature sequence is padded or clipped to a fixed length.

\subsection{Textual Feature Extraction}
In order to extract the feature representations for the article text, the pdf content is obtained through the Semantic Scholar Open Research pipeline (SSORP) \cite{lo-etal-2020-s2orc}. SSORP uses the SCIENCEPARSE\footnote{https://github.com/allenai/science-parse} and GROBID\footnote{https://github.com/kermitt2/grobid} APIs for text extraction from pdf. For the How2 dataset, the video transcriptions are manually annotated and transformed into a text feature set for training. In contrast, the acoustic features for \model\ are not transformed into the text as they are characterized by a variety of non-native English accents and a high error rate for speech-to-text models. Both the textual representations are tokenized using the vanilla BART tokenizer and transformed to word vectors using standard Transformer positional encoding embeddings.

\subsection{Dual-fused Hyper-complex Transformer}
We propose a  dual-fused hyper-complex Transformer (DFHC) for the task of multimodal text summarization. Compared to multi-head attention, the hyper-complex layer allows \model\ to efficiently capture the intricacies between different modalities and learn better representations \cite{https://doi.org/10.48550/arxiv.2102.08597} in the hyper-complex space. 
For the block DFHC, we represent the text as $X$ and pass it through the hyper-complex layer to extract the (Q) Query, (K) Key and (V) Value transformations as follows: $Q, K, V = \Phi(\text{HCL}(X))$, where $HCL(X) = Hx + b$. Here $H \in \R^{m \times n}$ is constructed by a sum of Kronecker products and is given by $H = \sum_{i=1}^n P_i \otimes Q_i$. The $P_i$ and $Q_i$ are the parameter matrices, and $\otimes$ represents the Kronecker product. 

The final attention score $A$ is obtained as:

\begin{equation*}
 A = \text{softmax}(\frac{Q K^T}{\sqrt{d_k}})
\end{equation*}

Here $Q$ represents the query value, $K$ represents the key value, and $d_k$ represents the key dimension.

The HCL layers share attention weights among the multi-head attention heads. The multi-head attention weights are concatenated and represented as
\begin{align*}
X = \text{HCL}([H_{1} + ... + H_{Num_{h}}])
\end{align*}
Here $Num_{h}$ represents the attention head. The final output obtained from the HCL layer is represented as:

\begin{align*}
Y = \text{HCL}(\text{ReLU}(\text{HCL}(X))),    \end{align*}
The transformation $Y$ is passed through a multi-head attention block and is fused with the visual embeddings using the cross-model attention as discussed in Section \ref{sec:fusionattention}.

\subsection{Wasserstein Riemannian Encoder Transformer}  
We base our idea from \citet{wang-wang-2019-riemannian} to repurpose the Wasserstein Riemannian Autoencoder to Wasserstein Riemannian Encoder Transformer (WRET) in the summarization setting. 

For an input $X$ and a manifold $M$, the Riemannian manifold is represented as $(M, G)$, where $G$ represents the Riemannian tensor unit. For two vectors $u$ and $v$ in the tangent space $T_{z}M$, the inner product is computed using $ \langle u, v\rangle_G = u^T G(z) v$. The Wasserstein block acts as a Variational Autoencoder. However, we extract the feature dimension from the last layer and feed it to the attention block.

The Wasserstein Autoencoder optimizes the cost between the target data distribution $A_{x}(x)$ and the predicted data distribution $B_{x}(x)$ using: 
\begin{align*}
Dist(A_X, B_G) = \inf_{Q(Z|X) \in {Q}} 
&  E_{P_X}E_{Q(Z|X)}[c(X, G(Z))] \\
& + \lambda MMD(Q_Z, P_Z)
\end{align*}
where $G$ is the generator function, $\lambda$ is a learnable metric, $c$ is the optimal cost, and $D_z$ is approximated between $B_G$ and  $Q_Z(z) = \int q(z|x)p(x)dx$ using the Maximum Mean Discrepancy (MMD) \cite{NIPS2016_5055cbf4}. The MMD is computed using

\begin{align*}
MMD_k(P_Z, Q_Z)
    & = ||\int_\mathcal{Z}k(z, \cdot)dP_Z - \int_\mathcal{Z} k(z, \cdot) dQ_Z||
\end{align*}
We formulate a RNF function $F = f_K \ldots f_1 $, and optimize the following RNF-Wasserstein function,

\begin{align*} \nonumber
Dist(A_X, B_G)
&= \inf_{Q(Z|X) \in \mathcal{Q}} {P_X} Q(Z|X)[c(X, {G(Z')})] \\
& \qquad + \lambda MMD(Q_{Z'}, P_{Z'}) \\
& + \alpha (KLD(q( z| x)||p( z)) - \sum \log |det \frac{\partial f'}{\partial z}|)
\end{align*}

where $Z'=F(Z)$, KLD is KL \cite{Joyce2011} divergence, and $p(z)$ represents the posterior probability distribution. The MMD term is approximated using the Gaussian kernel $k( z, z') = e^{-|| z- z'||^2}$. The term $G(Z')$ represents the reconstructed feature set, which is then passed to a feed-forward layer. The attention weights are computed for $G(Z')$ and fused with the audio feature using cross-modal attention as discussed in Section \ref{sec:fusionattention}.

\subsection{Cross-model Attention} \label{sec:fusionattention}

We fuse the text-video and text-audio features using cross-modal attention to align the attention distribution obtained from the last layer. The text feature set projects the Query $(Q)$ value, while the video and audio features project the key $(K)$ and value $(V)$, respectively. The obtained $Q$, $K$, and $V$ representations are passed through cross-modal attention, and the final encoder representation $E_s$ is obtained.

\begin{align*}
Q &= Z_t W_q;\  K = Z_v W_k;\ V = Z_v W_v \\
E_s &= \text{softmax}\left(\frac{X_\alpha W_{Q_\alpha} W_{K_\beta}^\top X_\beta^\top}{\sqrt{d_k}}\right) X_\beta W_{V_\beta}    
\end{align*}

The amalgamated representation, represented as $E_s$, is subsequently integrated with the multi-head attention mechanism of the BART decoder to yield the final summary.

To the best of our knowledge, the application of hyper-complex Transformer and Wasserstein Riemannian flow for abstractive text summarization has not been explored yet.

\section{Baselines}
We benchmark our proposed model against 20 baselines -- six extractive text summarization, eight abstractive text summarization,  two video-based abstractive text summarization, and four multimodal-input-based abstractive text summarization baselines. We briefly elaborate on them below.

\paragraph{{\bf Text-only extractive summarization:}}
(i) {\bf Lead-2:} The top 2 sentences of the source documents are marked as the generated summary and evaluated against the target summary.
    (ii) {\bf LexRank:} It \cite{Erkan_2004} represents the source document sentence as nodes of a graph and edges as a similarity measure. The edge weights are computed using the eigenvector centrality and the token frequency.
    (iii) {\bf TextRank:}  Similar to LexRank, it \cite{textrank2004} also represents the source document as a fully-connected graph. All edge weights are given a unit weight, and later a derived version of PageRank re-ranks the edge weights. 
    (iv) {\bf MMR (Maximal Marginal Loss):}  The redundancy between the sentences is computed by mapping the query to the sentence \cite{10.1145/3130348.3130369}. The relevant sentences are kept in a cluster and filtered based on the similarity ranking.
    (v) {\bf ICSISumm:}  The coverage of the sentence in the final summary is optimized using the linear optimization framework \cite{favre2008tac}. Given a summary length bound, integer linear programming (ILP) solvers try to maximize the global topic coverage.
    (vi) {\bf BERTExtrative:}  The task of summarization is transformed into a binary classification problem \cite{liu2019fine}. The sentences are classified into two classes representing whether the sentence is a part of the final summary or not.

\begin{table*}[!h]\centering
\caption{Performance benchmark over six text-only Extractive (Extr) baselines (Lead-2, Lexrank, TextRank, MMR, ICSISumm, and BERTExtractive), eight text-only Abstrative (Abst) baselines (Seq2Seq, Pointer Generator (PG), CopyTransformer, Longformer, BERT, BART, T5, and Pegasus), two video-only baselines (Action feature, and Action feature with RNN), and four Multimodal baselines (HA, FLORAL, MFFG, and ICAF) over the datasets -- \model\, and How2. The benchmarks are evaluated over the Quantitative metric -- Rouge (Rouge-1 (R1), Rouge-2 (R2), and Rouge-L (RL)), and Qualitative metric -- BERTScore (BERTSc.) and FEQA. }\label{tab:results}
\begin{tabular}{p{1.4cm}|l|rrrrr|rrrrrr}\toprule
\multirow{2}{*}{Type} &\multirow{2}{*}{System} &\multicolumn{5}{c|}{\data} &\multicolumn{5}{c}{How2} \\\cmidrule{3-12}
& &R1 &R2 &RL &BERTSc. &FEQA &R1 &R2 &RL &BERTSc. &FEQA \\\midrule
\multirow{6}{*}{Extr-text} &Lead-2 & 22.82 & 4.61 & 15.47 & 61.27 & 32.45 & 43.96 & 13.31 & 39.28 & 71.56 & 32.28 \\
&LexRank & 27.18 & 6.82 & 17.22 & 63.23 & 34.21 & 27.93 & 12.88 & 16.93 & 64.52 & 31.89 \\
&TextRank & 27.43 & 6.86  & 17.41 & 63.34 & 34.29 & 27.49 & 12.61 & 16.71 & 64.55 & 31.92 \\
&MMR & 29.54 & 8.19 & 18.84 & 64.59 & 35.67 & 28.24 & 13.12 & 17.86 & 64.87 & 31.98 \\
&ICSISumm & 31.57 & 9.52 & 19.42 & 65.84 & 36.14 & 28.53 & 13.44 & 17.93 & 65.14 & 32.16 \\
&BERTExtractive & 31.52 & 9.49 & 19.31 & 65.83 & 36.13 & 27.18 & 12.47 & 15.38 & 63.47 & 31.67 \\
\hline
\multirow{8}{*}{Abst-text} &Seq2Seq & 23.54 & 5.61 & 15.48 & 62.47 & 31.57 & 55.37 & 23.08 & 53.86 & 76.15 & 36.48 \\
&PG & 23.59 & 5.78 & 16.21 & 62.71 & 31.84 & 51.68 & 22.63 & 50.29 & 73.47 & 35.37 \\
&CopyTransformer & 25.63 & 7.82 & 18.54 & 63.11 & 37.86 & 52.94 & 23.25 & 50.26 & 73.58 & 35.43 \\
&Longformer & 21.37 & 6.47 & 15.12 & 61.05 & 32.14 & 49.24 & 21.39 & 47.41 & 72.39 & 35.28 \\
&BERT & 24.87 & 8.85 & 18.33 & 62.91 & 31.89 & 53.74 & 23.86 & 48.06 & 73.45 & 35.62 \\
&BART & 26.13 & 9.69 & 19.62 & 64.24 & 38.64 & 53.81 & 23.89 & 48.15 & 73.51 & 35.68 \\
&T5 & 25.87 & 9.24 & 18.63 & 64.13 & 38.45 & 53.21 & 22.51 & 47.48 & 73.42 & 35.65 \\
&Pegasus & 26.66 & 9.83 & 19.26 & 64.85 & 36.98 & 53.87 & 23.91 & 48.17 & 73.61 & 35.70 \\
\hline
\multirow{2}{*}{Video only} &Action features only & 26.38 & 6.47 & 15.37 & 62.48 & 30.41 & 45.24 & 24.42 & 38.47 & 69.74  & 31.28 \\
&RNN (Action features) & 26.73 & 6.51 & 15.75 & 63.14 & 31.35 & 48.27 & 27.74 & 46.37 & 72.32 & 35.11 \\
\hline
\multirow{5}{*}{Multimodal} &HA & 29.32 & 11.84 & 26.18 & 67.24 & 39.37 & 55.82 & 38.31 & 54.96 & 77.15 & 38.55  \\
&FLORAL & 31.69 & 13.54 & 31.55 & 69.56 & 41.19 & 56.84 & 39.86 & 56.93 & 79.84 & 39.14 \\
&MFFG & 33.19 & 18.88 & 33.28 & 71.54 & 43.13 & 61.49 & 44.61 & 57.21 & 80.16 & 41.59 \\
&ICAF & {\em 36.38} & {\em 20.54} & {\em 34.52} & {\em 73.94} & {\em 45.63} & {\em 63.84} & {\em 44.78} & {\em 58.24} & {\em 82.39} & {\em 42.86} \\\cline{2-12}
&\textbf{\model}  & \textbf{41.62} & \textbf{22.69} & \textbf{37.87} & \textbf{78.39} & \textbf{48.46} & \textbf{67.33} & \textbf{48.71} & \textbf{61.83} & \textbf{84.11} & \textbf{44.82}\\
\hline
$\Delta_{\model}$ & BEST &  \textcolor{blue}{$\uparrow 5.24$} &  \textcolor{blue}{$\uparrow 2.15$} &  \textcolor{blue}{$\uparrow 3.35$} &  \textcolor{blue}{$\uparrow 4.45$} &  \textcolor{blue}{$\uparrow 2.83$} &  \textcolor{blue}{$\uparrow 3.49$} &  \textcolor{blue}{$\uparrow 3.93$} &  \textcolor{blue}{$\uparrow 3.59$} &  \textcolor{blue}{$\uparrow 1.72$} &  \textcolor{blue}{$\uparrow 1.96$}  \\
\bottomrule
\end{tabular}
\vspace{-3mm}
\end{table*}

\paragraph{\bf Text-only abstractive summarization:}
    (i) {\bf Seq2Seq:} It \cite{nallapati2016abstractive} uses the standard RNN network for both encoder and decoder with a global attention mechanism.
    (ii) {\bf Pointer Generator (PG):} It \cite{see2017get} extends the Seq2Seq network with the addition of the Pointing mechanism. The Pointing mechanism allows the network to either generate tokens from the vocabulary or directly copy from the source document.
    (iii) {\bf CopyTransformer:}  It \cite{gehrmann2018bottom}  uses the standard Transformer network. Similar to PG, a random attention head acts as a pointing mechanism. It also leverages a content selection module to enrich the generated summary.
    (iv) {\bf Longformer:}  Unlike the standard Transformer \cite{vaswani2017attention}, Longformer \cite{Beltagy2020Longformer} uses linear attention to cater to the long source document. The computed attention weights are a combination of global and windowed attention. 
    (v) {\bf BERT:}  It \cite{devlin2018bert} is an encoder-only language model trained on the token masking technique. We fine-tune BERT over the text-only setting till convergence.
    (vi) {\bf BART:} It \cite{lewis2019bart} is an encoder-decoder-based language model pre-trained on the phrase masking technique. Similar to BERT, we fine-tune BART till convergence.
    (vii) {\bf T5:} It \cite{2020t5} considers all NLP downstream tasks as text-to-text problem. As text-to-text uses the same model architecture and loss throughout all NLP problems. We fine-tune t5-base on the \data\ training data.
    (viii) {\bf Pegasus:} Similar to T5, Pegasus \cite{zhang2020pegasus} is also an encoder-decoder based model. However, the self-supervision objective is to mask sentences rather than token masking, helping the model generate contextual sentences. This pre-training objective vastly helps text generation tasks like summarization.

\begin{table*}[!htp]\centering
    \caption{Ablation study to show the efficacy of each module of \model.}\label{tab:ablation}
    \begin{tabular}{l|rrrr|rrrr}\toprule
    \multirow{2}{*}{System} &\multicolumn{3}{c}{\data} & &\multicolumn{4}{c}{How2} \\\cmidrule{2-5}\cmidrule{6-9}
    &Rouge-1 &Rouge-2 &Rouge-L &BERTScore &Rouge-1 &Rouge-2 &Rouge-L &BERTScore \\\midrule
    Transformer  & 25.63 & 7.82 & 18.54 & 63.11 & 52.94 & 23.25 & 50.26 & 73.58  \\
    \quad + DFHC & 29.37 & 11.78 & 23.19 & 67.81 & 57.34 & 28.71 & 56.02 & 77.31 \\
    \quad + WRET & 34.52 & 14.82 & 26.54 & 72.06 & 61.12  &36.89 &58.1 & 81.44 \\
    \quad + DFHC \& WRET  & 37.34 & 18.32 & 32.49 & 74.58 &64.23 &42.61 &59.02 & 82.45 \\
     \model & 41.62 & 22.69 & 37.87 & 78.39 & 67.33 & 48.71 & 61.83 & 84.11  \\
    \bottomrule
    \end{tabular}
    \vspace{-3mm}
\end{table*}

\paragraph{\bf Video-only abstractive summarization:} 
    (i) {\bf  Action features only:} The video feature representations are trained over a convolution layer and passed through attention and an RNN-based decoder. 
    (ii) {\bf RNN (Action features):} The video features are trained over the convolution layer and  passed through attention and an RNN-based encoder. The latent representation is finally fused using the hierarchical attention and passed onto the RNN-based decoder to generate the final summary.

\paragraph{\bf Multimodal abstractive summarization:}
    (i) {\bf HA}: The work of \cite{libovicky-helcl-2017-attention} is repurposed for the multimodal summarization task. The visual and textual features are fused with hierarchical attention \cite{palaskar2019multimodal} to align features and capture more context while generating the summary. 
    (ii) {\bf  MFFG:}  It \cite{liu-etal-2020-multistage} introduces multistage fusion with forget gate. The encoder part uses a cross-attention-based fusion with forget gates. The decoder is assembled using a hierarchical fusion network to capture only the important concepts and forget redundant information.
    (iii) {\bf FLORAL:} It \cite{ATRI2021107152} proposes a Factorized Multimodal Transformer based language model consisting of guided attention and multimodal attention layer to align attention scores of each modality and use speech and OCR text to guide the generated summary.
    (iv) {\bf ICAF:} It \cite{https://doi.org/10.48550/arxiv.2108.05123} utilises recurrent and contrastive alignment to capture the relationship between the video and text. It makes use of contrastive loss to align modalities in the embedding space resulting in enriched aligned summaries. 

\section{Experiments}
We perform extensive ablations to evaluate the efficacy of the proposed modules of \model\ and individual modalities. We explore how text, visual and acoustic features perform separately and jointly over \data\ and How2.

\textbf{Evaluation measures:}
We evaluate the performance of \model\  using both quantitative metrics - Rouge-1, Rouge-2, and Rouge-L and qualitative metrics -- BERTScore and FEQA. Rouge measures the recall of unigrams (Rouge-1), bigrams (Rouge-2), and $n$-grams (Rouge-L) between the generated and target summaries. BERTScore assesses the similarity between the generated and target summaries in the embedding space through the cosine similarity of the BERT embeddings. FEQA, a question-answer generation metric, evaluates the quality of the generated summaries by determining the number of answers mapped to questions generated from the target summaries.

We further perform human evaluations\footnote{The human evaluations were performed by 15 individuals with sufficient  background in NLP, machine learning and deep learning. The participants were aged between 22-28 years.} In the evaluations, we benchmark the summaries over the following parameters --- Informativeness, Fluency, Coherence and Relevance (c.f. Supplementary, Section \ref{sec:appendixhuman}). We randomly sample $40$ instances from the test set and evaluate them against the target summaries. We perform human evaluations for BART (text-only), T5 (text-only), MFFG (multimodal), FLORAL (multimodal) and \model\ (multimodal).

\subsection{Training}
\label{sec:training_params}
The implementation of the \model\ model was carried out by utilizing the Pytorch 1.8.1 framework on an NVIDIA A6000 GPU equipped with 46 GB of dedicated memory and  CUDA-11 and cuDNN-7 libraries. The model was initialized with pre-trained BART language model weights for the encoder and decoder and fine-tuned on the summarization dataset. In the implementation, the loss computation was only performed over the target sequence in adherence to the encoder-decoder paradigm. The hyper-parameters used in both the pre-training and fine-tuning phases are detailed in Section \ref{sec:appendixexperi}, and Table \ref{tab:params} (Supplementary).

\begin{table*}[!htp]\centering \label{tab:sample}
\caption{Comparison of target summary with six models -- Extractive (ICSISumm), Abstractive (Pointer Generator (PG), BART, Pegasus) and multimodal (ICAF and \model) models.}\label{tab:sample}
\begin{tabular}{l p{14.5cm}}\toprule
Model & Output \\\midrule
Target & In this paper, we propose an adversarial multi-task learning framework, where the shared and private latent feature spaces donot interfere with each other. This task is achieved by introducing orthogonality constraints. \\\midrule\midrule
ICSISumm & To prevent the shared and private latent
feature spaces from interfering with each other, we
introduce two strategies: adversarial training and
orthogonality constraints. \\\midrule
PG & propose multi-task learning for the generative , propose latent feature for multi task learning where the shared knowledge regarded as off the self knowledge and trasferred to new task.  \\\midrule
BART &  In this paper, we conduct experiment on 16 tasks demonstrate the benefits and propose multi-task learning framework, The dataset are shared and latent feature spaces. the dataset is prone. \\\midrule
Pegasus & In this paper, we propose an multitask learning framework, where we conduct experiments on 16 text classification tasks. our model is off the shelf and donot interfere with each other.  \\\midrule
ICAF & we propose an adversarial multi-task framework, where we conduct experiments demostrating private feature space do not interefere with eachother. The model is regarded as off the shelf and transferred to new task. \\\midrule
\model & In this paper, we propose an multi-task framework, the shared and private latent feature spaces not interfere with each other. We conduct experiments on 16 text classification tasks. \\
\bottomrule
\end{tabular}
\end{table*}

\subsection{Quantitative Analysis}
Table \ref{tab:results} compares the performance of \model\ with that of its baselines across the How2 and \data\ datasets. Our results demonstrate that \model\ outperforms the best baseline, ICAF, with a Rouge-1 score of $41.62$ and a Rouge-L score of $37.87$, an improvement of $+5.24$  and $+3.35$, respectively. When benchmarked against the How2 dataset, \model\ exhibits superior results, obtaining Rouge-1 of $67.33$ and Rouge-L of $61.83$, outperforming the best baseline (ICAF) by $+3.49$ and $+3.59$ points, respectively. With respect to the best text-only abstractive baseline, Pegasus (Rouge-1 and Rouge-2) and BART (Rouge-L), \model\ shows an improvement of $+14.96$ Rouge-1, $+12.86$ Rouge-2, and $+18.25$ Rouge-L. Similarly, \model\ surpasses ICSISumm, the best extractive baseline, with an improvement of $+10.05$, $+13.17$, and $+18.45$ on  Rouge-1, Rouge-2 and Rouge-L, respectively.

We also perform ablations to study the efficacy of individual modalities and modules of \model. Table \ref{tab:singleresult} demonstrates the performance improvements obtained when all three modalities are fused, while Table \ref{tab:ablation} showcases the contribution of individual modules of \model. These results serve to affirm our hypothesis that models specifically designed for longer summarization sequences are inadequate in extreme summarization tasks and that the integration of multiple modalities with text modality enhances the quality of the generated summary.


\paragraph{\textbf{Congruency of multi-modalities}} The performance of various unimodal and multimodal text summarization systems is shown in Table \ref{tab:results}. The results demonstrate that for unimodal variants, the lead2, which was reported to be a strong baseline for datasets like CNN/Dailymail \cite{hermann2015teaching} and MultiNews \cite{Multinews2019}, fails to perform effectively, indicating that the latent structure of the scientific text is distinct, and the information has a heterogeneous distribution throughout the document. In a similar vein, the text-only abstractive baselines perform inadequately over both the How2 and \data\ datasets. On the other hand, the extractive baselines, which are able to identify the prominent sentences that start with ``we propose'' or ``we introduce'', perform better than the text-only abstractive baselines yet still provide only a limited context of the whole article. Meanwhile, the two video-only baselines display performance that is comparable to the best abstractive baselines, signifying that multimodal features do indeed contribute to generating more informative and coherent summaries. No baselines using audio-only features were run as audio captures only the amplitude shift and intonations, which do not constitute a sufficient feature set in the vector space. As indicated in Table \ref{tab:results}, the multimodal baselines outperform the text-only and video-only baselines. The HA model outperforms the best abstractive baseline by $+2.66$ Rouge-1 and $+6.92$ Rouge-L, demonstrating the significance of combining multimodal signals with text-only modalities. The fusion of video with text helps the model attain better context in the vector space, even the audio features aid in the mutual alignment of modalities leading to more diverse and coherent summaries. Evidently, all the remaining multimodal baselines show a remarkable improvement in performance over all the text-only extractive, text-only abstractive, and video-based baselines.


\begin{table}[!htp]\centering
\caption{Performance benchmark for each modality of \model.}\label{tab:singleresult}
 \small
\begin{tabular}{l|rrrr}\toprule
Modality &Rouge-1 &Rouge-2 &Rouge-L \\\midrule
Text +Audio & 27.46 & 7.47 & 19.62 \\
Audio +Video & 27.62 & 7.53 & 20.11 \\
Text +Video & 28.05 & 7.83 & 24.49 \\
Text +Audio +Video & 41.62 & 22.69 & 37.87 \\
\bottomrule
\end{tabular}
\vspace{-4mm}
\end{table}


\begin{table*}[!htp]\centering
\caption{Human evaluation scores over the metrics -- Informativeness (Infor.), Fluency, Coherence, and Relevance for the text-based baselines (BART and T5), multimodal baselines (MFFG, FLORAL, and \model) on the \model\ and How datasets.}\label{tab:humans}
\small
\begin{tabular}{l|r|rrrr|rrrrr}\toprule
\multirow{2}{*}{Modality} &\multirow{2}{*}{System} &\multicolumn{4}{c}{\data} &\multicolumn{4}{c}{How2} \\\cmidrule{3-10}
& &Infor. &Fluency &Coherence &Relevance &Infor. &Fluency &Coherence &Relevance \\\midrule
Abstractive-text &BART &2.81 &2.51 &2.94 &2.85 &2.34 &2.37 &2.46 &2.54 \\
Abstractive-text &T5 &2.78 &2.49 &2.81 &2.74 &2.33 & 2.28 & 2.43 & 2.54  \\
Multimodal &FLORAL & 3.2 & 3.03 & 3.02 & 3.11 & 3.13 & 3.14 & 3.08 &3.13 \\
Multimodal &MFFG &3.21 & 3.05 & 3.09 & 3.11 & 3.17 & 3.21 & 3.04 & 3.11 \\
Multimodal & \model & \textbf{3.46} & \textbf{3.32} & \textbf{3.27} & \textbf{3.29} & \textbf{3.34} & \textbf{3.27} & \textbf{3.21} & \textbf{3.18} \\
\bottomrule
\end{tabular}
\vspace{-3mm}
\end{table*}

\subsection{Qualitative Analysis}
We also conduct a qualitative evaluation of the generated summaries utilizing BERTScore and FEQA (c.f. Table \ref{tab:results}).
Both metrics use the text modality from the source and target to assess the quality. On the \data\ data, \model\ achieves  $78.39$ BERTScore and $48.46$ FEQA, surpassing the best baseline (ICAF) by $+4.45$ and $+2.83$ points, respectively. For the How2 dataset, \model\ obtains  $84.11$ BERTScore and $44.82$ FEQA, outperforming the best baseline (ICAF) by $+1.72$ and $+1.96$ points, respectively. Similar to the quantitative benchmarks, the multimodal baselines outperform the text-only extractive, abstractive, and video-only baselines by a substantial margin.

\subsection{Human Evaluation}
The quantitative enhancements are further reinforced by human assessments. As shown in Table \ref{tab:humans},  \model scores highest over the datasets -- \data\ and How2, demonstrating that the generated summaries are highly faithful, pertinent, and coherent in comparison to the other baselines. Although  \model\ demonstrates some deficiencies in the coherence criterion in the human evaluations, it still performs significantly better than the other baselines. A manual examination of the generated summaries and an analysis of the findings are presented in Section \ref{sec:erroranalysis}.

\subsection{Error Analysis} \label{sec:erroranalysis}
The limitations of extractive and abstractive baselines in generating extreme summaries are evident. Extractive systems rely on direct copying of phrases from the source document, often resulting in a single-line summary containing limited information diversity. This is reflected in their performance compared to abstractive text-only and a few multimodal (HA and FLORAL) baselines, as seen in Table \ref{tab:sample}. The text-only abstractive baselines like Seq2Seq and PG fail to extract the paper's main contributions, while Transformer based methods like Longformer, BERT, etc., struggle to summarize the contributions in very few lines.

However, \model\ stands out as it is able to condense the three key contributions of the source article into a single sentence, demonstrating superiority over the other baselines. A manual inspection of instances where \model failed to generate a good summary reveals that the cause was often due to noisy text modality extracted from the article pdf, leading to non-coherent connections between phrases. Further, the data noise in the video and audio modalities arising due to different aspect ratios of presentation and speaker and non-native English accent speakers adds to the perplexity of modality alignment.  


\section{Deployment - Continuous Human Feedback}
The performance improvements across the quantitative and qualitative metrics over the \data\ dataset motivated us to assess \model\ more rigorously. After controlled alpha testing of \model\ using human evaluations, we deployed \model\ as a web-based API (the technical details of API are discussed in Section \ref{sec:appendixdeploy} (Supplementary)) in an in-house tool. The API is currently hosted on a local server with an A6000 (48 GB) GPU, and the endpoints are accessible across the institute. For input, the API takes either a web URL consisting of direct links to the video and the article pdf or a separate file pointer to upload files directly from the local workstation. The current response time for the API is ~$2-3$ minutes, which is considerably high as a wait time. However, to cut down on the revert-back time, we cache all the responses to provide immediate output to the already processed queries. During the production stage, our aim will be to reduce the inference time to less than 1 minute. However, to achieve this, we will aim to  optimize the model by reducing model parameters and distilling it. Catering to the data regulations, we remove all the videos and article pdf's from our system after processing and only store the generated summaries and metadata for mapping queries to the cached data. The metadata includes the article title, author description, keywords, and month/year of publishing.  Further, we do not track users' identities nor store any user-specific information on our servers.


\section{Conclusion}
We introduced a novel task of extreme abstractive text summarization using multimodal inputs. 
we curated \data, a unique  large-scale dataset for extreme abstractive text summarization that encompasses videos, audio, and text, as well as both author-written and expert-annotated summaries. Subsequently, we introduced \model, a novel model that employs a dual fused hyper-complex Transformer and a Wasserstein Riemannian Encoder Transformer to efficiently capture the relationships between different modalities in a hyper-complex and latent geometric space. The hyper-complex Transformer captures the intrinsic properties between the modalities, while the Wasserstein Riemannian Encoder Transformer captures the latent structures of the modalities in the latent space geometry, enabling the model to generate diverse sentences. To assess the \model, we conducted thorough experiments on \data\ and How2 datasets and compared their performance with 20 baselines. Overall, \model\ demonstrated superior performance both qualitatively and quantitatively. 

\begin{acks}
The authors acknowledge the support of Infosys Centre for AI (CAI) at IIIT-Delhi and ihub-Anubhuti-iiitd Foundation set up under the NM-ICPS scheme of the DST. 
\end{acks}

\bibliographystyle{ACM-Reference-Format}
\balance
\bibliography{bibliography}


\appendix
\section{Experimentation and Deployment}\label{app:depl}
We discuss the experimentation environment and the deployment parameters in this section.
\subsection{Experimentation Details} \label{sec:appendixexperi}
We ran several sets of experiments over our proposed \model\ model to figure out the most optimal set of hyperparameters. Table \ref{tab:params} describes the most optimal hyperparameters for \model\ over the \data\ dataset. We initialize the \model\  weights using the pre-trained BART and then train the network further over the \data\ data. During training, we set the learning rate as $3e-5$ and set the lambda value as $18$. The gradients are accumulated for five iterations, and tri-gram blocking is used to penalize the decoder.

\begin{table}[!htp]\centering
\caption{HyperParameters used to train \model.}\label{tab:params}
    \begin{tabular}{l|rr}\hline
    \textbf{Parameter} &\textbf{Value} \\ \hline
    Epochs & 55 \\
    Accumulate gradient steps & 5 \\
    Ranking loss margin &0.001 \\
    MLE weight & 0.1 \\
    Warmup steps & 10000 \\
    Max learning rate & 3e-5 \\
    Max source length & 512 \\
    Training max summary length & 36 \\
    Testing max summary length & 40 \\
    Num of Beams & 4 \\
    GPU &2 X A6000 \\
    VMemory &48GB \\ \hline
    \bottomrule
    \end{tabular}
\end{table}

\begin{figure}[!h]
    \centering
    \scalebox{0.82}{
    \includegraphics[trim={0 1.5cm 1.3cm 0cm},clip]{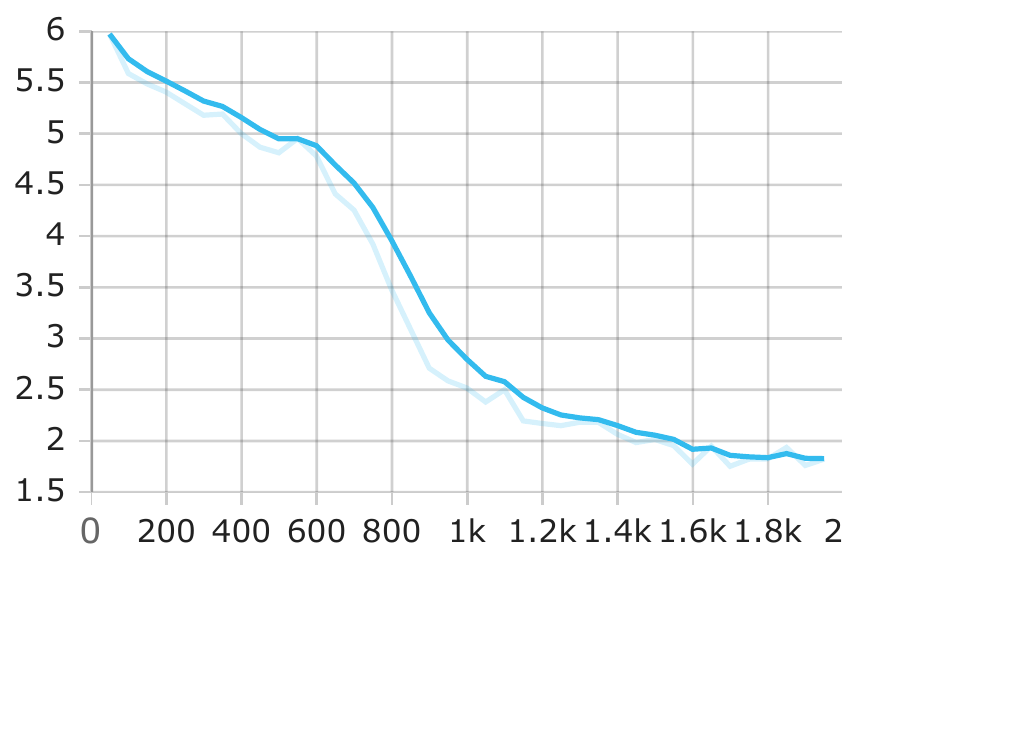}}
    \caption{Loss momentum of \model\ over the \data\ dataset.}
    \label{fig:loss}
\end{figure}

We train the model till loss converges and the validation accuracy does not improve for five continuous iterations. Figure \ref{fig:loss} shows the variation of loss till $2000$ iterations.

\subsection{Deployment} \label{sec:appendixdeploy}
The web API for \model\ has been created using the Flask \footnote{https://flask.palletsprojects.com/en/2.2.x/} framework. As Flask only allows single-user access at a given time, the API is running on top of the Gunicorn \footnote{https://gunicorn.org/} framework, allowing multiple users to access the API at the same time. At a particular instance, the API is able to handle a load of four concurrent requests without major variation in inference time. The trained \model\ model is hosted separately as an API running on Docker, giving the users option to either generate TLDR summaries using the web interface or directly by calling the \model\ using any programming language.

\section{Human Evaluation Setup} \label{sec:appendixhuman}
\begin{figure*}[!h] \label{fig:formh}
    \centering
    \scalebox{0.5}{
    \includegraphics[trim={0 0cm 0cm 0cm},clip]{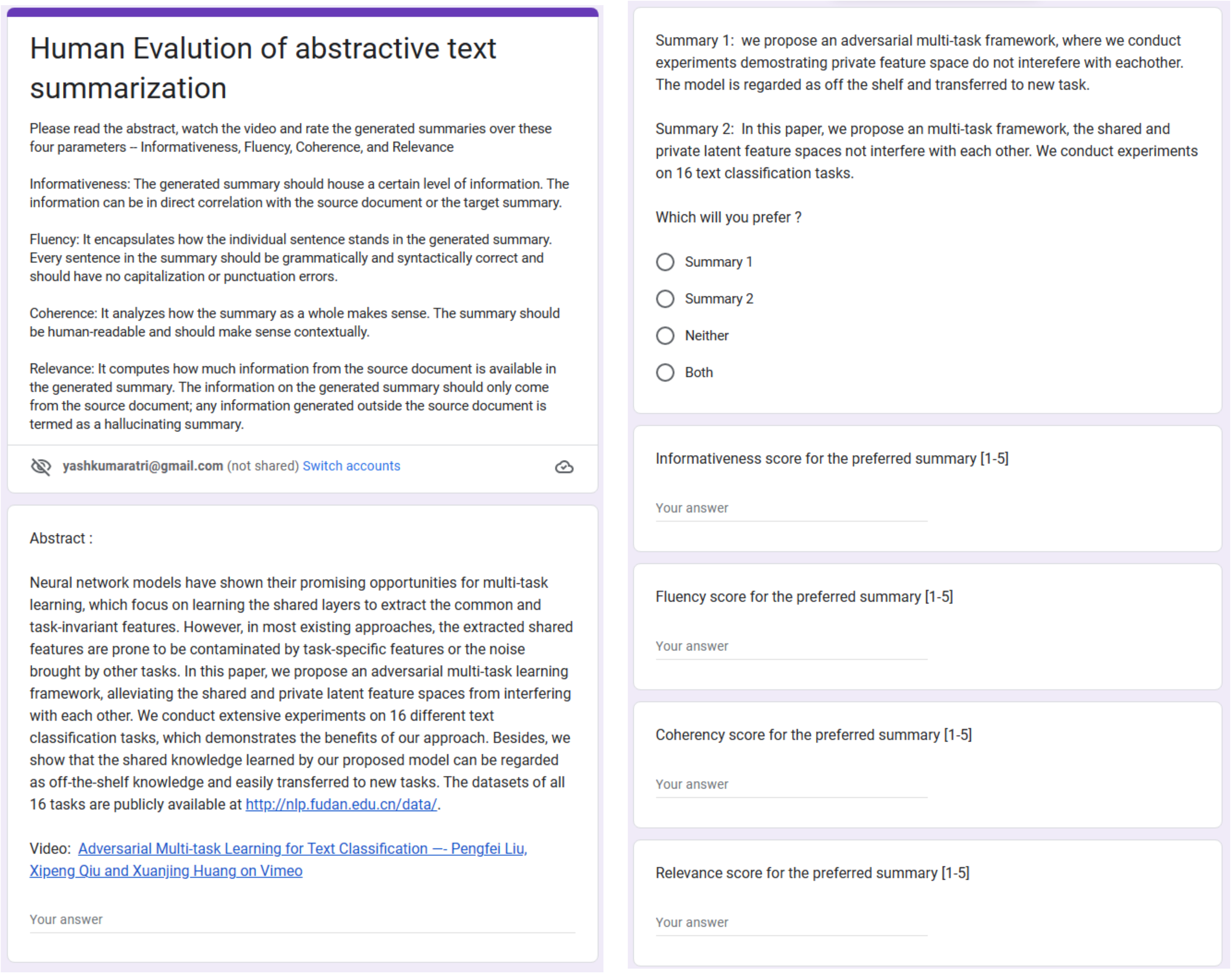}}
    \caption{Human Evaluation form for collection feedback over the \model\ and baseline generated summaries.}
\end{figure*}

We evaluate the generated summaries over four \cite{Multinews2019} parameters -- Informativeness, Fluency, Coherence and Relevance. Figure \ref{fig:formh} shows the form utilised by human evaluators to benchmark the generated summaries against the competing baselines.
\begin{enumerate}
    \item Informativeness: The generated summary should house a certain level of information. The information can be in direct correlation with the source document or the target summary.
    \item Fluency: It encapsulates how the individual sentence stands in the generated summary. Every sentence in the summary should be grammatically and syntactically correct and should have no capitalization or punctuation errors.  
    \item Coherence: It analyzes how the summary as a whole makes sense. The summary should be human-readable and should make sense contextually. 
    \item Relevance: It computes how much information from the source document is available in the generated summary. The information on the generated summary should only come from the source document; any information generated outside the source document is termed as a hallucinating summary.
\end{enumerate}




\end{document}